\documentclass[letterpaper, 10 pt, conference]{ieeeconf}  

\IEEEoverridecommandlockouts                              

\usepackage{graphicx}
\usepackage{hyperref}
\usepackage{caption}
\usepackage{subcaption}
\usepackage{multirow}
\usepackage[T1]{fontenc}
\usepackage{array}
\usepackage{float}
\usepackage{textcomp}
\usepackage{amsmath}
\usepackage{amssymb}
\usepackage{stackrel}
\usepackage{lipsum}
\usepackage{algorithmic}
\usepackage{color}

\title{\LARGE \bf
Robotic Cane as a Soft SuperLimb for Elderly Sit-to-Stand Assistance*
}
\author{Xia Wu$^{1}$, Haiyuan Liu$^{1}$, Ziqi Liu$^{1}$, Mingdong Chen$^{1}$, Fang Wan$^{2}$, \\
Chenglong Fu$^{1}$, Harry Asada$^{3}$, Zheng Wang$^{1}$, and Chaoyang Song$^{1,*}$
    \thanks{*This work was supported by Southern University of Science and Technology, National Students' Innovation and Entrepreneurship Training Program (No. 201914325006), and SUSTech-MIT Joint Center for Mechanical Engineering Education and Research.}
\thanks{$^{1}$Xia Wu, Haiyuan Liu, Ziqi Liu, Mingdong Chen, Chenglong Fu, Zheng Wang and Chaoyang Song are with Department of Mechanical and Energy Engineering, Southern University of Science and Technology, 
        Shenzhen, Guangdong 518055, China. 
        {\tt\small \{11612026, 11510872, 11712526, 11711903\}@mail.sustech.edu.cn, \{fucl, wangz, songcy\}@sustech.edu.cn}}%
\thanks{$^{2}$Fang Wan is a Visiting Scholar with the SUSTech Institute of Robotics, Southern University of Science and Technology, 
        Shenzhen, Guangdong 518055, China. 
        {\tt\small sophie.fwan@gmail.com}}%
\thanks{$^{3}$Harry Asada is with the Department of Mechanical Engineering, MIT, 
        Cambridge, MA 02138, USA. 
        {\tt\small asada@mit.edu}}%
\thanks{$^{*}$Chaoyang Song is the corresponding author.}%
}
\begin{document}
\maketitle
\thispagestyle{empty}
\pagestyle{empty}
\begin{abstract}
    Many researchers have identified robotics as a potential solution to the aging population faced by many developed and developing countries. If so, how should we address the cognitive acceptance and ambient control of elderly assistive robots through design? In this paper, we proposed an explorative design of an ambient SuperLimb (Supernumerary Robotic Limb) system that involves a pneumatically-driven robotic cane for at-home motion assistance, an inflatable vest for compliant human-robot interaction, and a depth sensor for ambient intention detection. The proposed system aims at providing active assistance during the sit-to-stand transition for at-home usage by the elderly at the bedside, in the chair, and on the toilet. We proposed a modified biomechanical model with a linear cane robot for closed-loop control implementation. We validated the design feasibility of the proposed ambient SuperLimb system including the biomechanical model, our result showed the advantages in reducing lower limb efforts and elderly fall risks, yet the detection accuracy using depth sensing and adjustments on the model still require further research in the future. Nevertheless, we summarized empirical guidelines to support the ambient design of elderly-assistive SuperLimb systems for lower limb functional augmentation. 
\end{abstract}
\section{Introduction}
\label{sec:Introduction}
    Low fertility rates and progressive lengthening of life expectancy keep exacerbating the population aging problem among the developed and even developing countries \cite{WB:2019}. A growing demand emerges for the need for elderly care for those who have difficulties in daily activities. Literature survey shows that social robots as a companion provide a positive influence with reduced stress and an improved sense of security among people with aging and dementia \cite{Alonso2019Social}. Robotic solutions for the elderly need to take both physical and cognitive consideration into the design process of intelligent and interactive agents \cite{Cynthia2019Designing}. While the elderly are becoming more open to social robotic solutions, it remains a challenge to integrate novel solutions such as wearable technologies with active assistance for elderly users.
    
    The sit-to-stand movement is closely related to the falls of the elderly, which is the leading cause of injury and the primary etiology of accidental deaths among the elderly over 65 years of age, which accounts for 70 percent of accidental deaths in persons 75 years of age and older \cite{Tinetti2003Preventing}. The sit-to-stand test is a widely adopted metric for activities of daily lives (ADLs) to measure the necessary mobility skills of the elderly \cite{Podsiadlo1991Timed, Beauchet2011Timed}. It is commonly practiced when trying to leave the seat, such as the bed, chair, or toilet, during the transition between sitting and walking \cite{Isaacs1985Clinical}.
    
    \begin{figure}[tbp]
        \centering
        \textsf{\includegraphics[width=1\columnwidth]{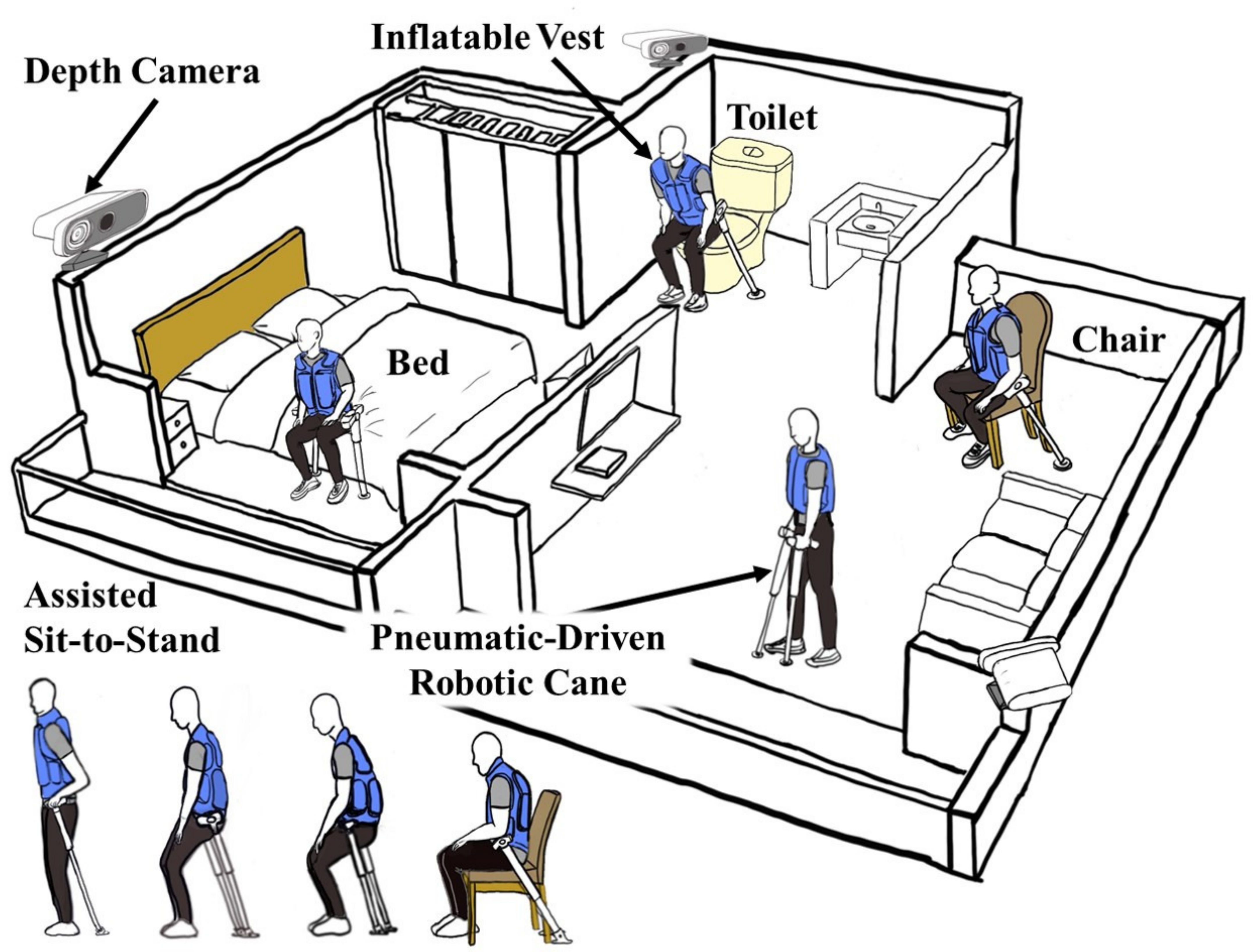}}
        \caption{Overview of the proposed soft superlimb system including a pneumatically-driven robotic cane, an inflatable soft vest, and a privacy-safe depth camera for sit-to-stand assistance at home, including bed, chair, and toilet.}
        \label{fig:PaperOverview}
    \end{figure}
    
    The supernumerary robotic limbs (SuperLimbs) emerge in recent research exploring the cognitive boundary of brain-machine interaction for wearable robotics \cite{Asada2017Independent}, \cite{Gonzalez2018Design}. Much of the SuperLimbs research focuses on healthy subjects with normal brain and motor functions, with SuperLimbs as an augmentation system for collaborative assistance in certain work conditions \cite{Gonzalez2019Loop}. However, there remains a research gap exploring the potential daily assistance of SuperLimbs for elderly users with degraded brain and motor functions in the ADLs, which is the focus of this paper. In particular, we propose the design of a robotic cane system with pneumatic actuation, inflatable vest, and privacy-safe range sensing to explore the use of SuperLimbs for elderly assistive care during the sit-to-stand transition at home, shown in Fig. \ref{fig:PaperOverview}.

\subsection{Related Work}
    \subsubsection{The Sit-to-Stand Motion Transition} 
    The literature classifies the sit-to-stand movement into three phases, namely trunk flexion, hip lift-off, and the knee-hip extension \cite{Millington1992Biomechanical}. Trunk flexion results in forward shifting, transforming the support to only feet \cite{Hughes1994Chair}. Hip lift-off is the key event in transforming the forward motion in an upward direction. After that, the maximum vertical ground reaction force occurs. Knee-hip extension lifts the body weight in the vertical direction until full standing \cite{Scarborough2007Chair}. 

    \subsubsection{Robot Design for the Elderly}
    Existing literature on the robot design guidelines for the elderly is mainly focused on social robots for cognitive care, such as stress relief and mental support \cite{Broekens2009Assistive}. While it is generally accepted that robot designs should take the user needs of the elderly into considerations, yet the task is still more complicated in practice \cite{Pape2002Shaping}. The adoption of agents of artificial intelligence is considered as a potential solution to address the individual characteristics of the elderly user \cite{Cynthia2019Designing}, but further developed is required to integrate these agents with active physical support for elderly usage.
    
    \subsubsection{Motion Detection for Elderly Care}
    The traditional elderly care mainly depends on human labor, which is costly. The researchers have been searching for effective methods to offer autonomous and round-the-clock caring for the growing aging communities. Some wearable devices installed with biosensors and inertial sensors have been developed for the elderly health-monitoring \cite{pantelopoulos2009survey} and activity recognition \cite{avci2010activity}. However, these intrusive wearable devices need to be worn all day round. Due to the limitations of measuring range, users may need to wear multiple sensors when a large range of daily activities are monitored \cite{chen2017survey}. 
    
    On the other hand, the researchers have been actively searching for non-intrusive approaches to detect the daily life activities with minimal effect to the users. RGB-video based action recognition provides an economical solution \cite{foroughi2008intelligent, rougier2011fall}. However, since this method requires the RGB-video sequence to be recorded, it has the potential of violating the user’s privacy. With the emergence of a low-cost depth sensor, the researchers have begun to utilize depth information for elderly activity detection \cite{presti20163d}. Depth cameras, in particular, provide cost-effective real-time 3D body movement data. Without recording the raw depth-image, depth cameras, which only records the desired body joint data, is privacy-preserving, comparing to the previous approaches. Furthermore, the vision system indicates the potential to empower the intelligent around-the-clock ambient interaction in the elderly healthcare with little labor cost, as demonstrated in \cite{YeungA} and \cite{3d_point_cloud_icu_care}. With the unprecedented development of 3D technology, the depth-sensor will be able to detect various activities and serves for other purposes, such as fall prevention and long-term health monitoring. 

\subsection{Proposed Method and Contributions}
    In this paper, we propose a soft robotic technique to design a robotic cane system, providing sit-to-stand assistance for the elderly usage at home. Through the integration of a pneumatically-driven linear actuator for active lower-limb support, an inflatable vest for compliant interaction between the human body and the robotic cane, and a privacy-safe depth vision for intention detection to compensate for potential issues with cognitive usage by the elderly. The contributions of this paper are listed as the following.
    \begin{itemize}
        \item The design and development of a robotic cane system as a supernumerary robotic limb for the elderly assistance;
        \item The integration of soft and non-intrusive robotic techniques for active elderly assistance in forms of pneumatic actuation, inflatable vest, and privacy-safe sensing;
        \item A modified biomechanical model of sit-to-stand movement with an active cane for analysis and control with experimental verification;
        \item The experimental validation of each component of the proposed robotic cane system;
        \item Preliminary design guidelines of supernumerary robotic limb system for elderly care. 
    \end{itemize}
    
    The rest of this paper is organized as follows. Section \ref{sec:Method} formulates the problem of sit-to-stand transition, followed by the proposal of a modified biomechanical model with a robotic cane attached. In this section, we also introduce the system design of the supernumerary robotic limb system of the robotic cane, inflatable vest, and privacy-safe camera. Section \ref{sec:ExpResults} presents the experiment setup, procedure, and results to validate the feasibility of such a robotic cane for sit-to-stand motion transition. Section \ref{sec:Discussion} includes the discussion of the experiment results on the soft hardware design for the elderly, as well as the design guidelines for supernumerary robotic limbs for the elderly care. The conclusions and limitations of this paper are enclosed in the final section, which ends the paper.

\section{Robotic Cane as Soft SuperLimbs for Elderly}
\label{sec:Method}
\subsection{Problem Formulation \& Biomechanical Modeling}
\begin{figure}[htbp]
        \centering
        \textsf{\includegraphics[width=1\columnwidth]{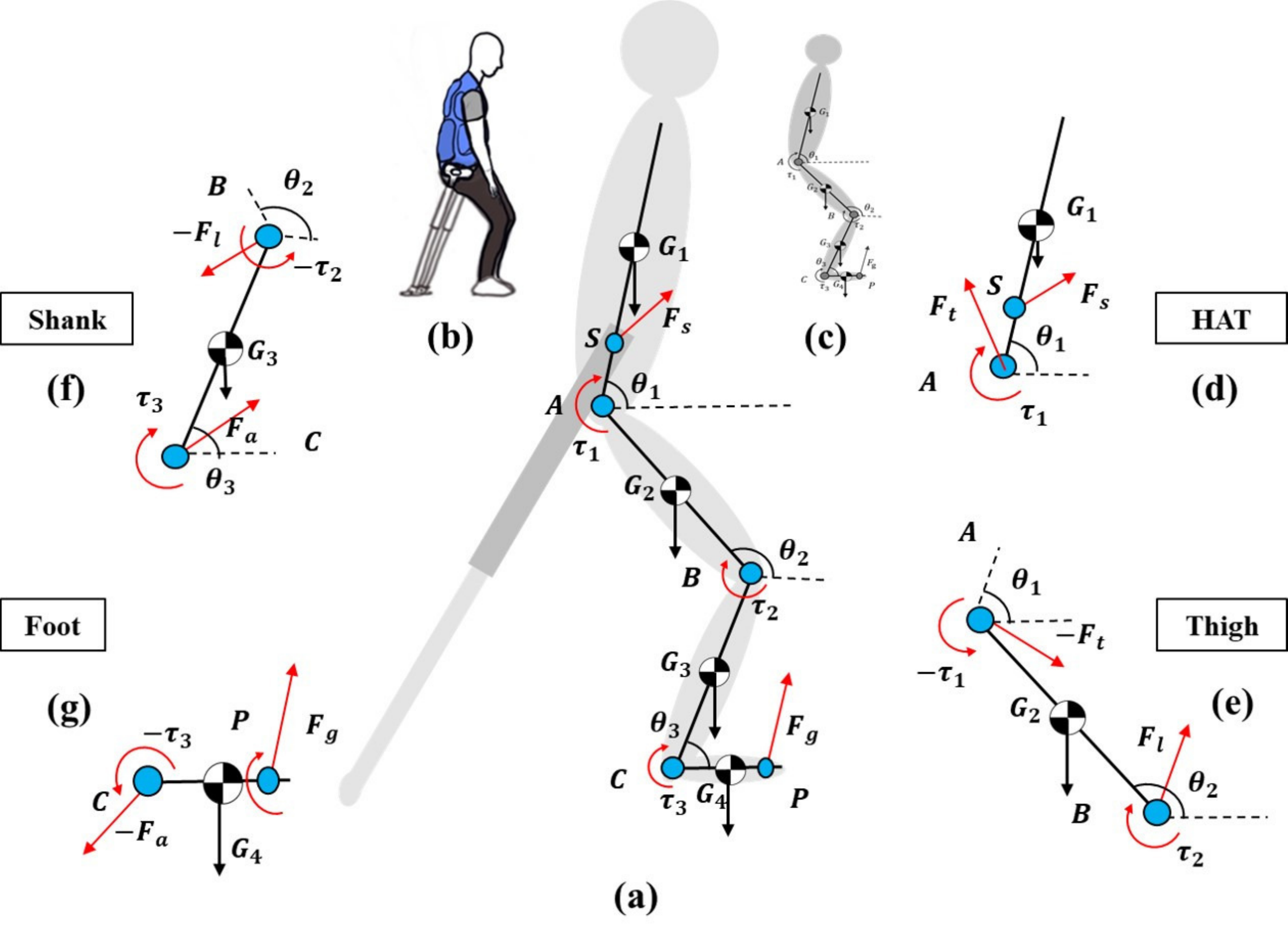}}
        \caption{The modified biomechanical model of sit-to-stand movement (a) and the free-body diagrams of HAT (d), thigh (e), shank (f), and foot (g) segments. The conceptual design with the cane is shown on the left (b) while the classical sit-so-stand model on the right (c).}
        \label{fig:ModifiedModel}
    \end{figure}

    In order to better adjust the output force and control travel distance of the robotic cane during the sit-to-stand process, it is important to formulate the biomechanical model. The proposal of an active robotic cane with linear motion modifies the original model with an extra ``leg''. As shown in Fig. \ref{fig:ModifiedModel}, the human is modeled as a kinematic chain of four-linked rigid mechanism with three degrees of freedom in the sagittal plane. The body segments included are feet, shanks, thighs, and head-arms-trunk as a rigid bar named HAT \cite{Sibella2003Obese}. Anthropocentric values are calculated based on \cite{Leva1996Adjustments}. The sit-to-stand movement of the elderly is slow, and the acceleration is relatively small compared to the weight.
    
    Here are the assumptions for the biomechanical model: 1) The sit-to-stand movement symmetric in frontal and horizontal plane; 2) The feet fix to the ground during the whole movement; 3) Inertial of lower limbs not significant.
    
    The external force is anchored at the body, noted as $F_s$ fixed at $S$ point. In the process of natural stand-up, $F_s$ is zero. Point $S$ varies with the real situation. Therefore, the case that the subject stands up and stands assisted actively by the robotic cane system ware both described in one model, indicated as Fig. \ref{fig:ModifiedModel}.
    
    For the HAT segment, 
    \begin{equation}
        \tau_{1} + \overrightarrow{AG_{1}}\cdot\overrightarrow{G_{1}} + \overrightarrow{AG_{1}}\cdot{(m_{1}\cdot{\overrightarrow{a}})} - J_{1}\cdot{\ddot{\alpha}} + \overrightarrow{AS}\cdot\overrightarrow{F_{s}} = 0
        \label{eq:EyeOnBase}
    \end{equation}
    \begin{equation}
        \overrightarrow{F_{s}} + \overrightarrow{G_{1}} + \overrightarrow{F_{t}} = m_{1}\cdot\overrightarrow{a}
        \label{eq:EyeOnBase}
    \end{equation}
    where $\tau_{1}$ is the unknown hip moment, $\overrightarrow{G_{1}}$ is the mass vector of the HAT, $\overrightarrow{AG_{1}}$ is the vector from the hip to the HAT mass center. $m_{1}\cdot{\overrightarrow{a}}$ is the inertial vector, $J_{1}\cdot{\ddot{\alpha}}$ is the mass inertial moment with respect to the perpendicular from the sagittal plane, both calculated from the sensing data. $\overrightarrow{AS}$ is the vector pointing from the hip to the location of support force contact point and $\overrightarrow{F_{s}}$ is the support force on $S$. By Newton's second law, the force from thigh $\overrightarrow{F_{t}}$ is calculated.
    
    For the thigh segment, 
    \begin{equation}
        \tau_{2} + (-\tau_{1}) + \overrightarrow{BG_{2}}\cdot\overrightarrow{G_{2}} + \overrightarrow{BA}\cdot(\overrightarrow{-F_{t}}) = 0
        \label{eq:EyeOnBase}
        \end{equation}
        \begin{equation}
        \overrightarrow{F_{l}} + \overrightarrow{G_{2}} + \overrightarrow{-F_{t}} = 0 
        \label{eq:EyeOnBase}
    \end{equation}
    where $\tau_{2}$ is the unknown knee moment, $\overrightarrow{G_{2}}$ is the mass vector of the thigh, $\overrightarrow{BG_{2}}$ is the vector from the knee to the thigh mass center, $\overrightarrow{BA}$ is the vector from the knee to the hip. $\overrightarrow{F_{l}}$ is the unknown reaction force from the shank.
    
    For the shank segment,
    \begin{equation}
        \tau_{3} + (-\tau_{2}) + \overrightarrow{CG_{3}}\cdot\overrightarrow{G_{3}} + \overrightarrow{CB}\cdot(\overrightarrow{-F_{l}}) = 0
        \label{eq:EyeOnBase}
        \end{equation}
        \begin{equation}
        \overrightarrow{F_{a}} + \overrightarrow{G_{3}} + \overrightarrow{-F_{l}} = 0
        \label{eq:EyeOnBase}
    \end{equation}
    where $\tau_{3}$ is the unknown ankle moment, $\overrightarrow{G_{3}}$ is the mass vector of the shank $\overrightarrow{CG_{3}}$ is the vector from the ankle to the shank mass center. $\overrightarrow{CB}$ is the vector from the ankle to the knee. $\overrightarrow{F_{a}}$ is the unknown reaction force from the foot.
    
    For the foot segment,
    \begin{equation}
        -\tau_{3} + \overrightarrow{CG_{4}}\cdot\overrightarrow{G_{4}} + \overrightarrow{CP}\cdot\overrightarrow{F_{g}} + M= 0
        \label{eq:EyeOnBase}
    \end{equation}
    where $\overrightarrow{G_{4}}$ is the mass vector of the foot, $\overrightarrow{CG_{4}}$ is the vector from the ankle to the foot mass center. $P$ is the pressure center, $M$ is the total moment from the force plate, $\overrightarrow{CP}\cdot\overrightarrow{F_{g}}$, denoted as $M_{c}$, is the moment by ground reaction force $F_{g}$. $M_{c}$ is the key component that needs to be compared.

\subsection{System Overview of Robotic Cane as a SuperLimb}
\begin{figure}[bp]
        \centering
        \textsf{\includegraphics[width=1\columnwidth]{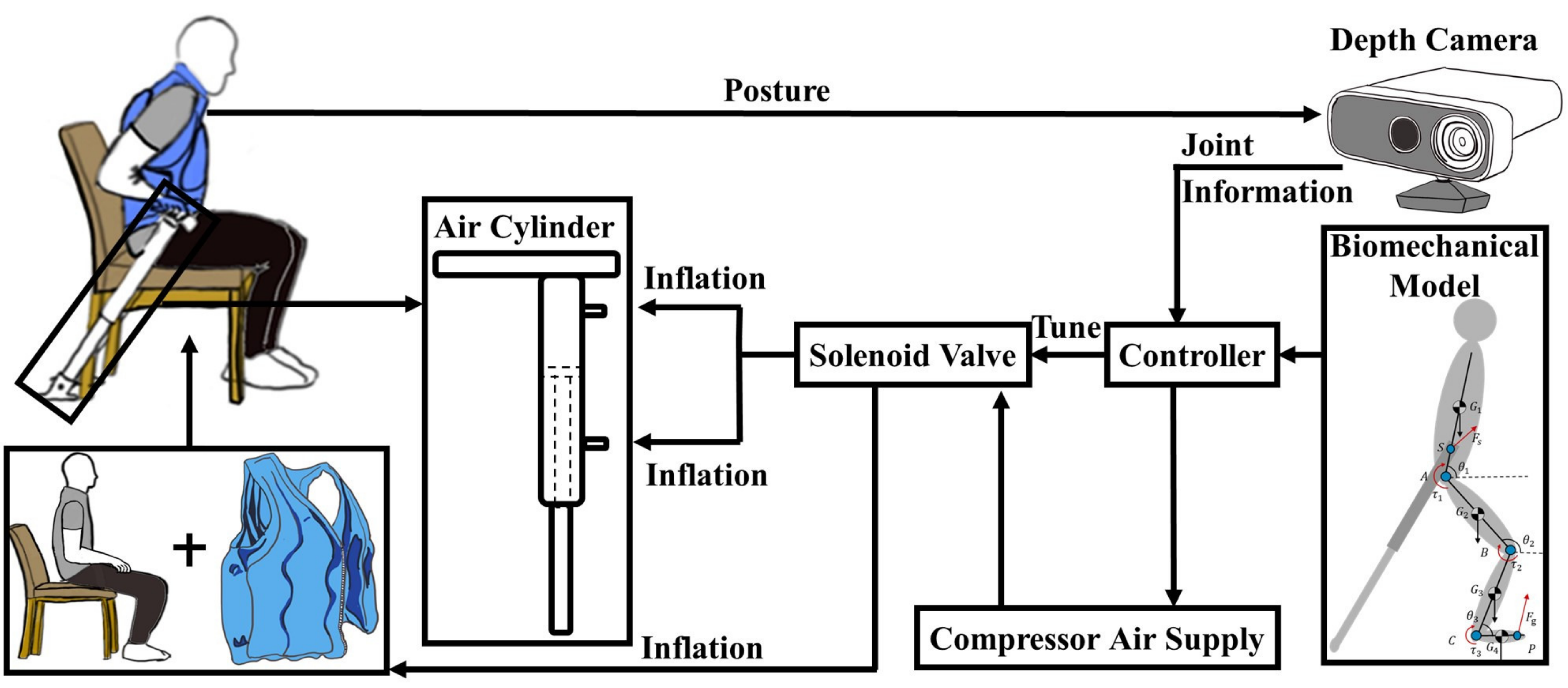}}
        \caption{Proposed control algorithm for the system.}
        \label{fig:RoboticCane-ControlSystem}
    \end{figure}

    The proposed SuperLimb system is illustrated in Fig. \ref{fig:RoboticCane-ControlSystem}, where soft and non-intrusive techniques are adopted to accommodate the elderly need in ADLs. A pneumatically-driven linear cylinder is used for lower limb support, which provides a reasonably large and compliant thrust force within a safe operating pressure. The overall shape of the pneumatic cylinder is redesigned as a cane that is commonly used by the elderly for body weight support. Considering the reaction force from the cylinder to the human body during physical support, such force may be too much for the elderly to carry. Therefore, an inflatable vest is added to the system, which can be a regular vest with customized design patterns on the outside, as long as individual air-bag arrays are embedded within to provide distributed support during motion. The depth camera is used to collect depth images that are privacy-safe even at home environment to detect the intention from the cane users during motion transition. Data collected from the depth cameras are then transmitted to the controllers of the robotic cane and inflatable vest, so that the elderly users may not need to worry about the operational control of the valves during motion transition, reducing the stress and complexities in manual control and valve adjustment. 
    
    The robotic cane is to be used in the home environment at places such as the bedside, toilet, or the chairs in the living room, as shown in Fig. \ref{fig:PaperOverview}. The assumption is that the elderly user will have the vest put on at first. Then, when moving towards the edge of the seat with the robotic cane in hand, the depth camera will capture the movement and detect the intentions after that. The processed data on user intention will then be transmitted to the on-board controller of the robotic cane to feed the modified biomechanical model with a robotic cane. Prior model tuning is required during the initial installation of the whole system to adjust the individual need of the elderly user. Once a sit-to-stand intention is detected, the inflatable vest will be pressurized, and then the robotic cane will start to move, providing axial thrust to assist the elderly user in completing the sit-to-stand process. 
        
    \begin{figure}[htbp]
        \centering
        \textsf{\includegraphics[width=1\columnwidth]{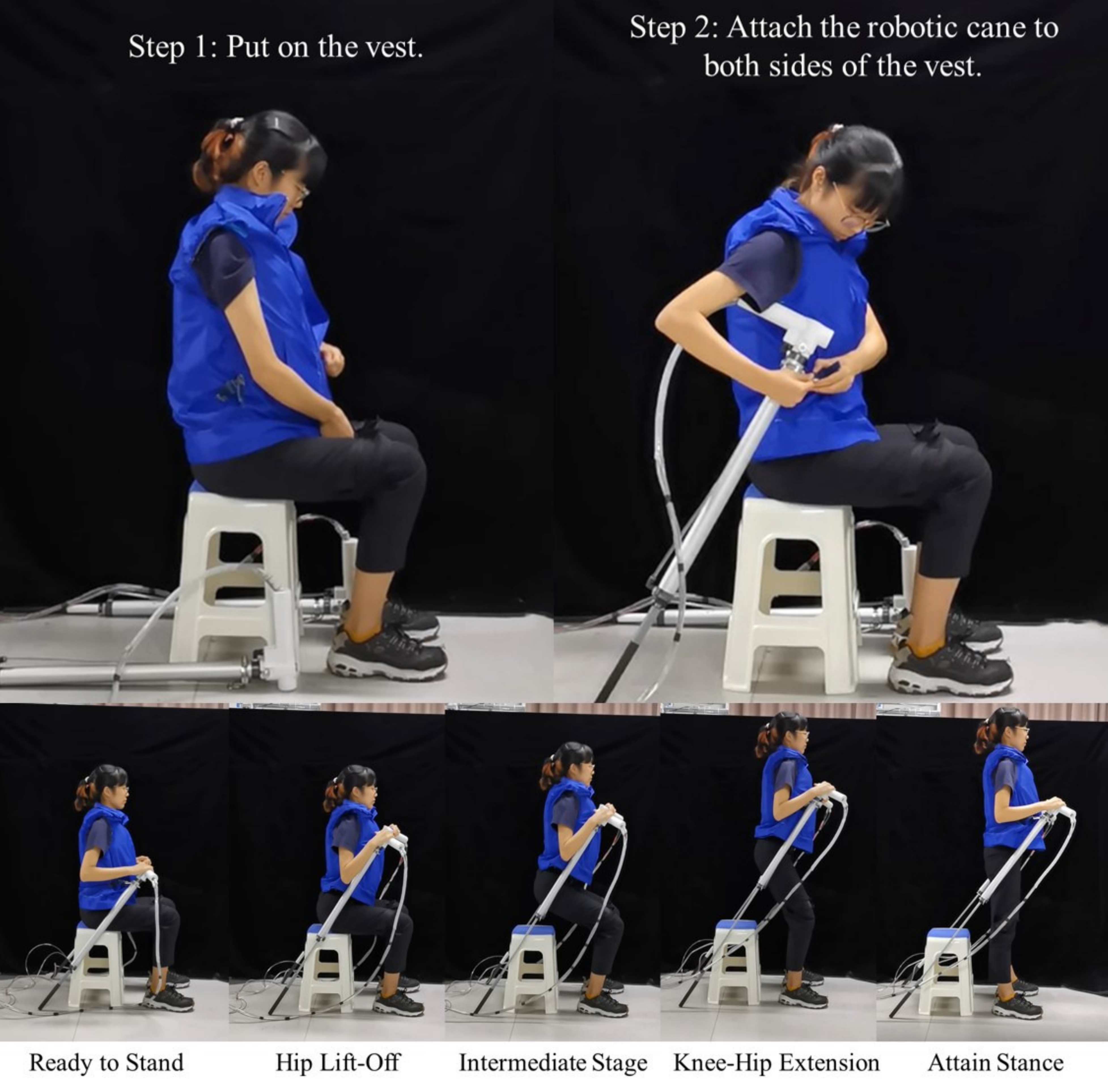}}
        \caption{A two-step setup and demonstration of the robotic cane system in chair. Here, the operator use hand to actuate the robotic cane by pushing a button on the handle instead of using hands for upper body support.}
        \label{fig:SitToStand}
    \end{figure}
    
    The adoption of ``soft'' hardware is a crucial design feature of the overall system. The pneumatically-driven cylinder in the robotic cane aims at providing a gradual and compliant, yet powerful, thrust while providing lower limb support. The inflatable vest is used as a soft interface between the cane and the wearer, where a flexible joint will be used to connect the vest with the cane. In this way, the thrust force provided by the cane will be distributed through the vest to act on the wearer to avoid impact or hard push. The depth-camera is a much less intrusive sensing solution, providing more detailed information to detect whole-body user intention. After providing the assistance, the vest will deflate, and the cane will be adjusted to a locked position, so that the elderly user may use the robotic cane usually as a cane for walking assistance. 
     
\subsection{Design of the Robotic Cane}
    Assistive cane or walking stick is a common type of crutch device used by the elderly for weight redistribution, improved stability, and tactile information about the ground \cite{Kaye2000Mobility}. Ten percent of US adults over 65 years use a cane\cite{Bradley2011Geriatric}. However, there remains a limited design to robotize the cane for elderly usage. A potential drawback for a robotic cane would be an increase in weight and size, which may cause discomfort by the elderly users. Engineering design issues on this problem may be solved by using more advanced material and a light-weight power source. In this paper, we propose a pneumatically-driven robotic cane to explore the potential design issues related to sit-to-stand assistance for the elderly usage. 
    
    Before establishing a proof-of-concept prototype, we should first consider requirements especially length, travel distance, and bearing capacity. Mean height and mean weight of elderly men are 173.4$cm$ and 88.3$kg$, used as references \cite{MeanWeight2018}. The upper limb accounts for approximate 53\% of total height \cite{Leva1996Adjustments}. With the assumption that hip height change before and after sit-to-stand is  equal to the thigh length, the travel distance is roughly 30\% \cite{Leva1996Adjustments}. Consequently, the major design requirements can be summarized as Table \ref{tab:designRequirements}. 
    
    \begin{table}[htbp]
        \caption{Major design requirements for the robotic cane.}
        \label{tab:designRequirements}
        \begin{center}
            \textsf{\includegraphics[width=1\columnwidth]{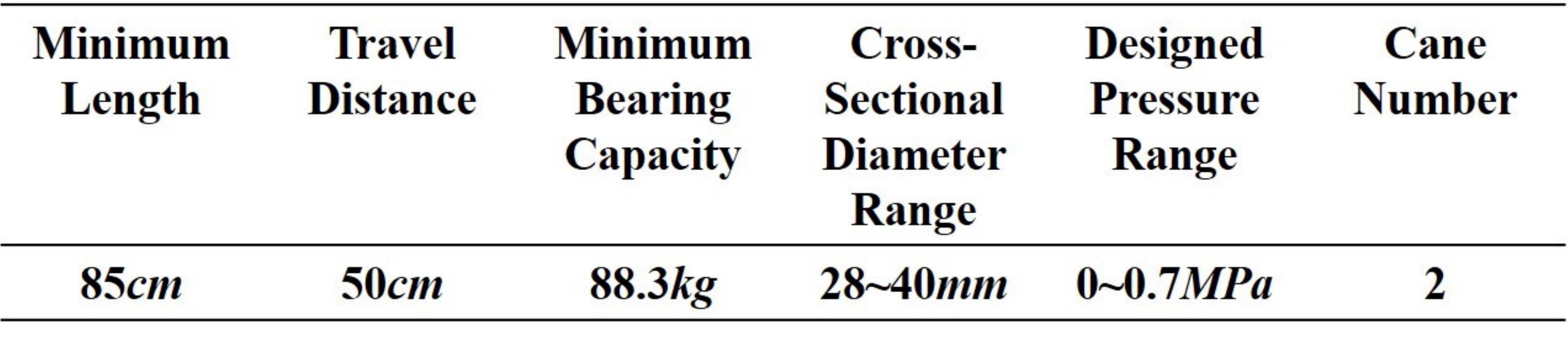}}
        \end{center}
    \end{table}

\begin{figure}[tbp]
        \centering
        \textsf{\includegraphics[width=1\columnwidth]{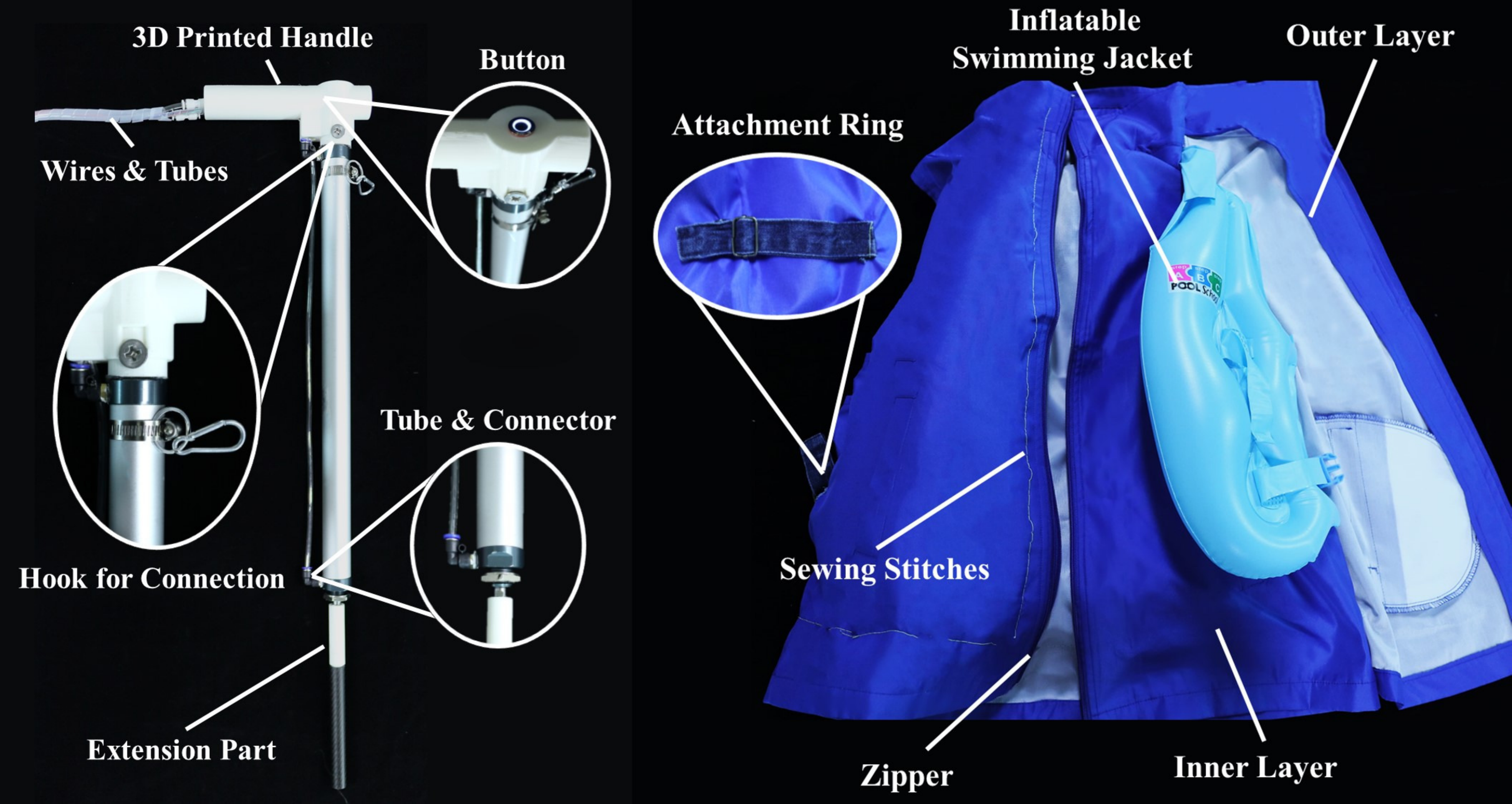}}
        \caption{The proof-of-concept prototype: a redesigned air cylinder as the robotic cane and a swimming jacket enclosed in the vest.}
        \label{fig:RoboticCane-MechDesign}
    \end{figure}

    We modified a pneumatic cylinder of 32$mm$ cross-sectional diameter with 500$mm$ travel distance and 0.8$Mpa$ maximum pressure by adding a handle on top with controllers and connectors inside, as shown in Fig. \ref{fig:RoboticCane-MechDesign}(a). The maximum thrust force of a single cylinder is about 644$N$, meaning that a double-cane design can provide a lifting force for an object of about 130$kg$, which satisfy the requirement. As our intentional use is for assistive support, a single cane design is sufficiently powerful to help an elderly with reasonably motor functions for sit-to-stand purposes. 

     A 3D printed T-shape handle is designed and assembled to the air cylinder. The current version is still manually actuated, and a button is placed on top of the handle to operate the valve. The L-shaped air connectors reroute the tubes along the cylinder. The wires from the button together with tubes go through the handle to the back for connection. The electromagnetic valve controlled by the button is to switch the inlet tube. The hook under the handle is useful during attachment to the vest. Since the cane is not rigidly attached to the body, there remains certain flexibility from the cane when providing the thrust force to adjust to the human posture during sit-to-stand.
     
\subsection{Elderly Wearable using an Inflatable Vest}
     An inflatable vest is also added to the system to distribute the thrust force provided by the robotic cane evenly. The outer and inner layers of the vest are regular clothes, while the middle layer is air inflatable. As a proof-of-concept, an inflatable swimming jacket is sewed to the inside of the vest. When deflated, the vest is almost the same as a regular vest. Once inflated, the vest will pop up and conform to the shape of the body, forming a tight fit between the body and the vest. Attachment rings are sewed on to the side of the vest under the arm, above the waist line, which are to be connected with the hook on the cane during usage. When the cane is pressurized, the cane will push the rings on the vest, and the thrust force will be evenly distributed through the inflatable swimming jacket inside to achieve compliant support. As a result, the vest is able to hold the human body while the thrust force from the cane increases.
     
\subsection{Privacy-safe System for Intention Detection}
    The pattern of human motion can be described by the temporal and spatial movement of their skeleton. The development of the 3D depth camera has made it possible to trace human body skeleton joints economically. Microsoft Kinect for Azure released its body tracing software development kit \cite{shotton2013real-time}, by which the temporal-spatial positions of 32 human body skeleton joints are available. For privacy consideration, only the 3D position of each joint is recorded, rather than RGB-video. Firstly, the body skeleton joints are projected on the sagittal plane. The angles of ankle, knee, and hip are then calculated. As shown in Fig. \ref{fig:RoboticCane-ControlSystem}, by importing the joint information of the current state, the controller is able to calculate the error using the biomechanical model and adjust the robotic cane output force.
     
\section{Experiments}
\label{sec:ExpResults}
    Two experiments are designed to investigate (1) the feasibility of the robotic cane and inflatable vest to provide sit-to-stand assistance; (2) the rationality of the biomechanical model; and (3) the posture recognition during sit-to-stand against motion capture system for intention detection.

\subsection{Experiment Setup}
\begin{figure}[htbp]
        \centering
        \textsf{\includegraphics[width=1\columnwidth]{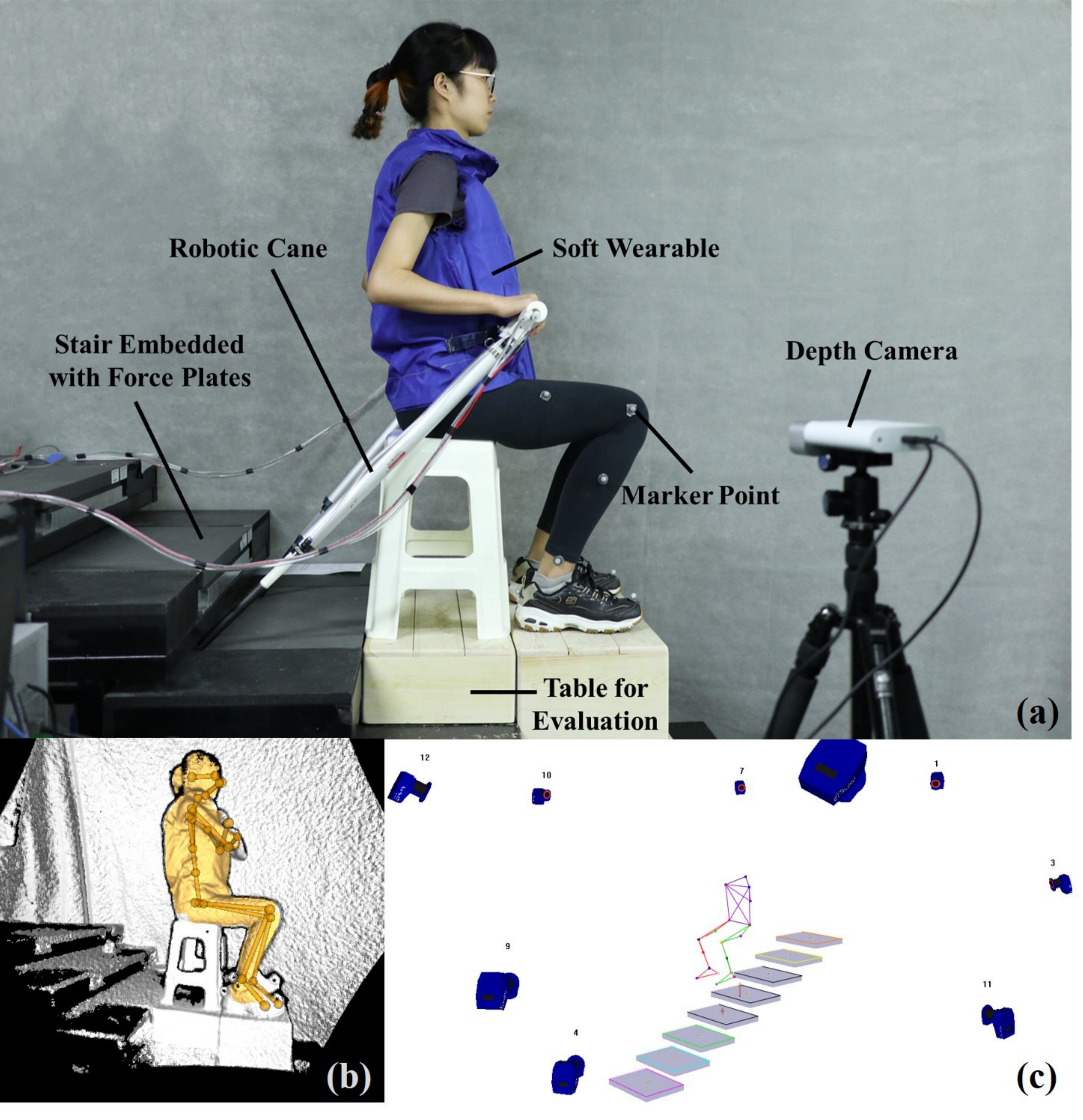}}
        \caption{Experiment setup: (a) for the experimental group assisted by the robotic cane while (b) and (c) for the standard control group, the data from the depth camera and motion capture system, respectively.}
        \label{fig:ExpSetup}
    \end{figure}

    Eight force plates (1200Hz, Bertec) were embedded in each step of a stair. In this experiment, at most three force plates were needed. Therefore, two tables were manufactured for elevation so that we could have a flat plane with three sets of independent force data collected, as shown in Fig \ref{fig:ExpSetup}(a). A depth camera (60Hz, Kinect for Azure) and a  twelve-camera motion capture system (120Hz, Motion Capture) recorded the whole process synchronously, as shown in Fig \ref{fig:ExpSetup} (b), (c). The chair with 40cm in height was used in the experiments.  
    
\subsection{Experiment Procedure}
    Each subject was required to finish two trials of the experiments, standard control group and experimental group. In the standard control group, the subject was expected to perform sit-to-stand with arms folded over the chest and without any external assistance. While in the experimental group, the subject stood up with robotic cane system assisted. Initially, the subject was seated on the chair on the middle force plate. The feet were placed on the front in parallel while the cane placed on the back one in the cane experiment. Nothing else contributed to the applied forces. In the initial position, the subject’s buttocks should be right at the back of the chair seat. Hands did not push off when starting. The first head and trunk orientations were not controlled. Subjects did not change the feet position once the initial state was established. 

    Each trail of experiments was repeated six times. For the control group, the rising process should follow the beat of a metronome, 60 beats per minute. The purpose was to restrict two trails of experiments within the similar time frame. Subjects were able to rest if needed. Force plates, the depth camera, and the motion capture system recorded the experiments synchronously.

\subsection{Results}
\begin{figure}[htbp]
        \centering
        \textsf{\includegraphics[width=1\columnwidth]{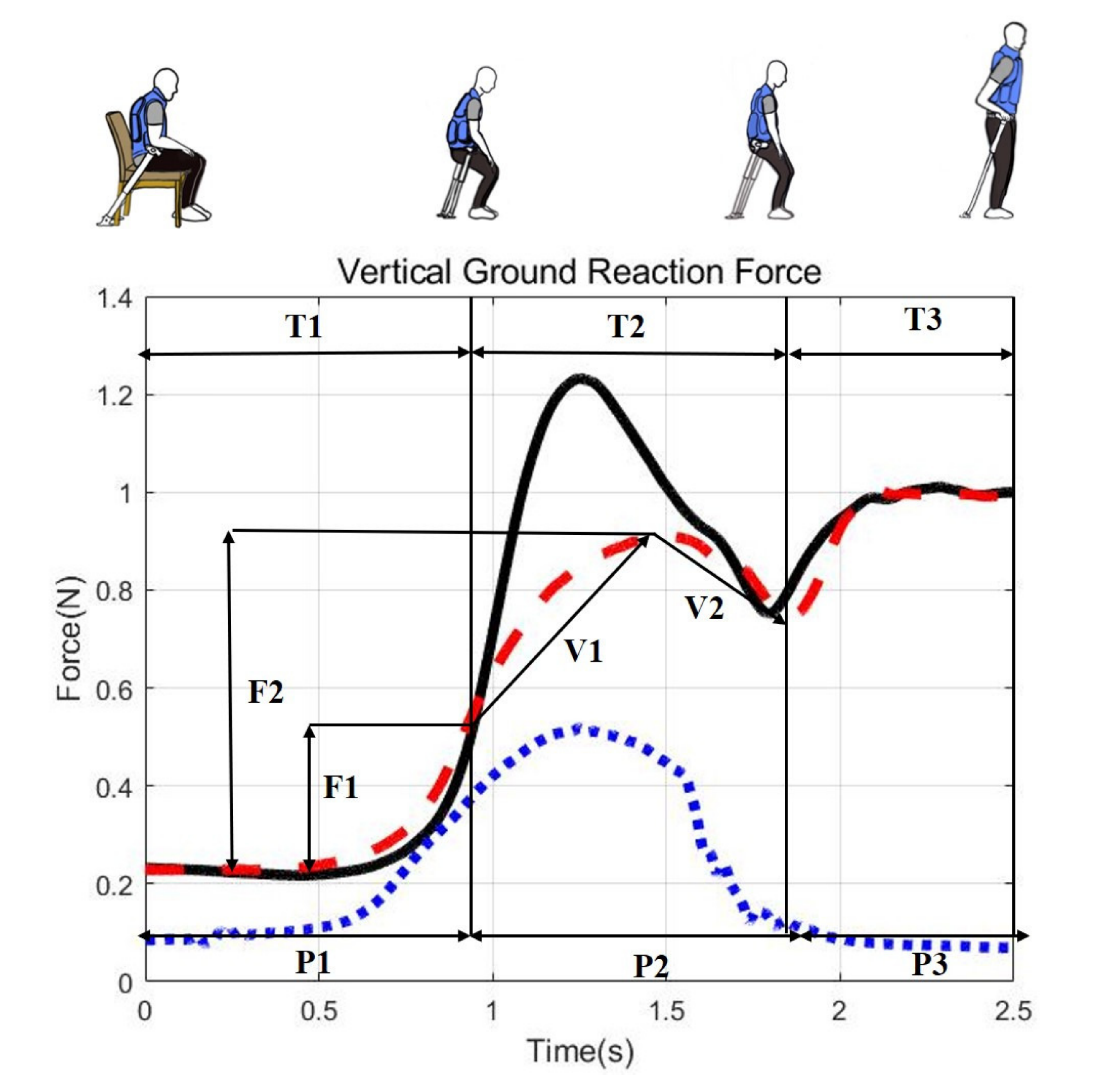}}
        \caption{Mean vertical ground reaction force and ten parameters during the sit-to-stand movement. The black solid line was the force on human feet in the standard control group while the red dash line using the robotic cane system. The blue dotted line was the force on the robotic cane.}
        \label{fig:ExpResult}
    \end{figure}

    Fig. \ref{fig:ExpResult} shows the mean vertical ground reaction force on each trial. More detailed values, mean and standard deviations, of ten ground reaction force parameters are shown in Table \ref{tab:ExperimentResult}, which characterized the process associated with falling risks of the elderly \cite{Yamada2005Parameters, Yamada2009Relationships}. $F1$ was the force magnitude at hip lift-off, while $F2$ was the peak ground reaction force. $T1$ started from beginning to hip lift-off, $T2$ started from hip lift-off to peak reaction force and $T3$ started from peak reaction force to completion of the sit-to-stand movement. $P1$, $P2$, $P3$ were impulse during $T1$, $T2$, $T3$, respectively. The last two parameters were velocity, describing the change rate of ground reaction force, $V1$ between hip lift-off to peak ground reaction force, and $V2$ during knee-hip extension.
    
\begin{figure}[htbp]
        \centering
        \textsf{\includegraphics[width=1\columnwidth]{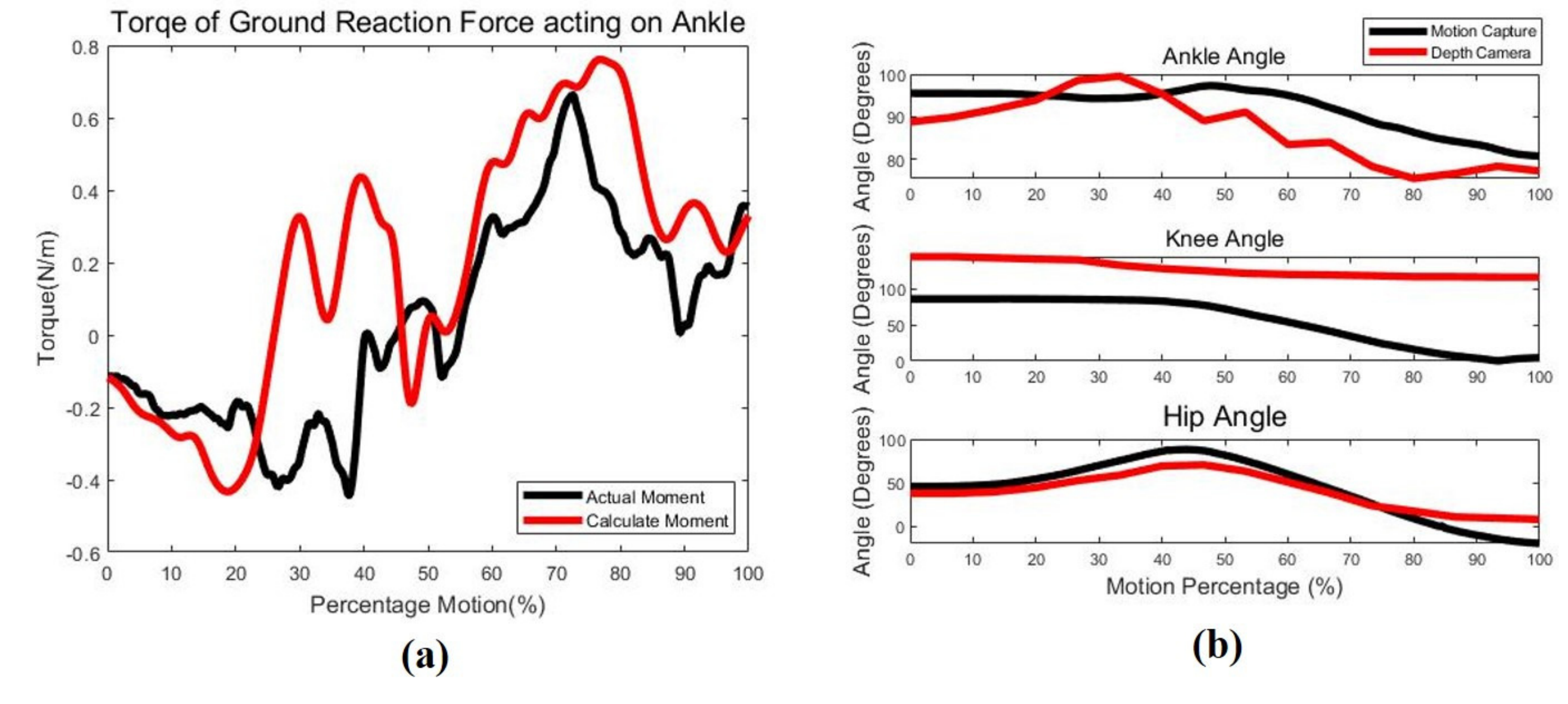}}
        \caption{Experiment results of (a) the biomechanical model and (b) the identification accuracy of the depth camera. The inaccuracy is potentially caused by mismatch in joint models on the angle and knee.}
        \label{fig:ModelCompare}
    \end{figure}

    Fig. \ref{fig:ModelCompare} (a) shows the torque acting on the ankle of real ground reaction force and calculation result from the biomechanical model.  Fig. \ref{fig:ModelCompare} (b) shows the detected joint angles from the depth camera against the motion capture system. The black lines were both ground truth values, while the red lines were calculated or detected results.
    
    \begin{table}[htbp]
        \caption{Mean values and standard deviations in ground reaction force of two experimental groups.}
        \label{tab:ExperimentResult}
        \begin{center}
            \textsf{\includegraphics[width=1\columnwidth]{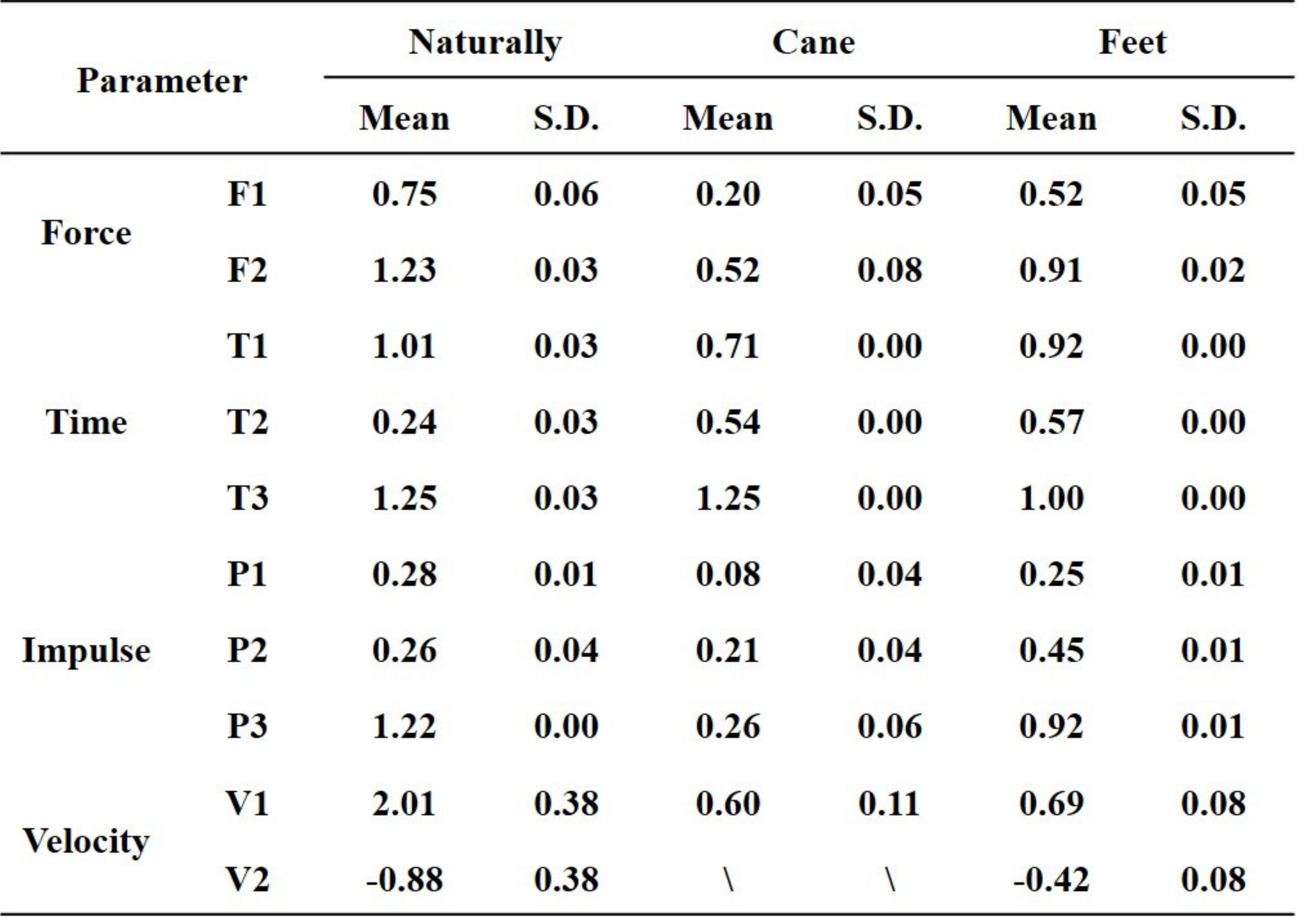}}
        \end{center}
    \end{table}
    
\section{Discussion}
\label{sec:Discussion}
\subsection{``Soft'' Hardware for the Elderly Care}
    \subsubsection{The Inflatable Vest as the Wearable} 
    In the proposed system, the soft wearable, inflatable vest is a key component, which builds a bridge between the human and robot. Instead of popular applications as actuation, here we present "support" for the soft robot application. It works to support the operator's upper body with comfort as well as safety and prevent stress concentration considering narrow contact area like a belt. Besides, a two-step setup is as simple as wear a normal jacket for the elderly, indicated in Fig \ref{fig:SitToStand}. 
      
    However, what we are curious about is that besides supporting, what else can the wearable do? The depth camera observes and detects the sit-to-stand movement from a side view currently. At least three cameras are needed in different locations of a home environment. The wearable gives an alternative for sensor integration. Used every day, the wearable, which is embedded with an inertial measurement unit (IMU) and a camera in the front, may be one of the simple ways for sensing. No other device is required for extra wear. The IMU is capable of body posture detection, while the camera "sees" the environment, considering bedside, chair, and toilet, which is level 2 in Fig. \ref{fig:ThreeLevelGuidlines}, a different design level from the proposed system. Apart from enabling the robot to sense the human intention, the wearable serves as a feedback intermediate between the robot and the human. Disney research put forward a force jacket that used airbags for feel effects in the virtual reality entertainment \cite{Effectofmusclestrength}. Inspired by it, some of the inflatable parts of the wearable can be used as feedback signals. Thus, both forward and backward sensing from the human to the robot are achievable.

    \subsubsection{The Pneumatic Cylinder as the Robotic Cane} 
    Fluidically driven, the robotic cane is another critical hardware in the system. The sit-to-stand mechanism assisted by the robotic cane can be analyzed by comparing three sets of ground reaction force data in Fig. \ref{fig:ExpResult} and Table \ref{tab:ExperimentResult}. For the human-robot interaction in the experimental group, there was a noticeable time delay in the motion between the robotic cane and the subject. The robotic cane moved first and gradually transmitted the force to the human body. Thus, the subject took longer both in $T1$ and $T2$ than the robotic cane. Before the subject reached the peak value of $F2$, the force on cane had started to decrease. Almost when the subject started to perform knee-hip extension, indicated as the valley of red dash line, the force on cane decreased to its weight. 
    
    The advantages of using the robotic cane system are fewer efforts for lower limbs and lower fall risk when we compare the force on subject of standard control group and experimental group. Force, $F1$ and $F2$, and velocity absolute values of the operator, $V1$ and $V2$, are good reflection. The experimental group showed smaller values of all the mentioned parameters than the standard control group. Thus, the force curve using the robotic cane was smoother and flatter. More importantly, the overshoot showing in the control group disappeared in the experimental one, which means that the robotic cane successfully compensates the portion of the force larger than the body weight. Thus less muscle strength and fewer efforts are required by the operator shortly after the hip lift-off phase. We can also conclude from $F2$ of the robotic cane that an external force of 52\% of body weight is sufficient for sit-to-stand assistance. Hip lift-off and Knee-hip extension phases are useful in predicting falling risks \cite{Sittostandfromprogressively}. The shorter these phases cost, the easier for the elderly to keep balance when standing up. When standing up assisted by the robotic cane, both $T1$ and $T3$ had smaller values than that of the control group. Therefore, the robotic cane shows good performances in supporting the user and reducing potential fall risks.
    
    \subsubsection{The Biomechanical Model with Depth Camera as a Non-Intrusive Sensor} 
    In order to verify the theoretical model, the torques of actual and calculated ground reaction force from the biomechanical model acting on the ankle were compared. From Fig. \ref{fig:ModelCompare} we can tell that there was a phase shift during the hip lift-off phase and some differences in the torque magnitude due to the assumptions for simplification, which were not exactly true. One possible reason was that the ankle position shook a lot during each cycle and its error could not be neglected compared to its change within centimeter range. Another possible explanation was that the error of each phase was independent from the others. Consequently, to improve the model, further calibration on with standard posture is needed for error reducing. 
    
    We used the Kinect Azure SDK to detect the human skeleton information in the sit-to-stand action. The algorithm performed well when the range of the movement was big while the detection precision was not good enough when the range of the movement is small. Since the change in the ankle data was relatively small, within 10 degrees in a cycle, little disturbance and error would lead to the oscillation and detection damages. Many other approaches based on deep learning can offer a robust understanding of the human activities \cite{YeungA}. It is possible to use the biomechanical model to retrieve the human activities from the depth images adjusting the output force of the robotic cane.
        
\subsection{Design Guidelines for the SuperLimbs}
\begin{figure}[htbp]
        \centering
        \textsf{\includegraphics[width=1\columnwidth]{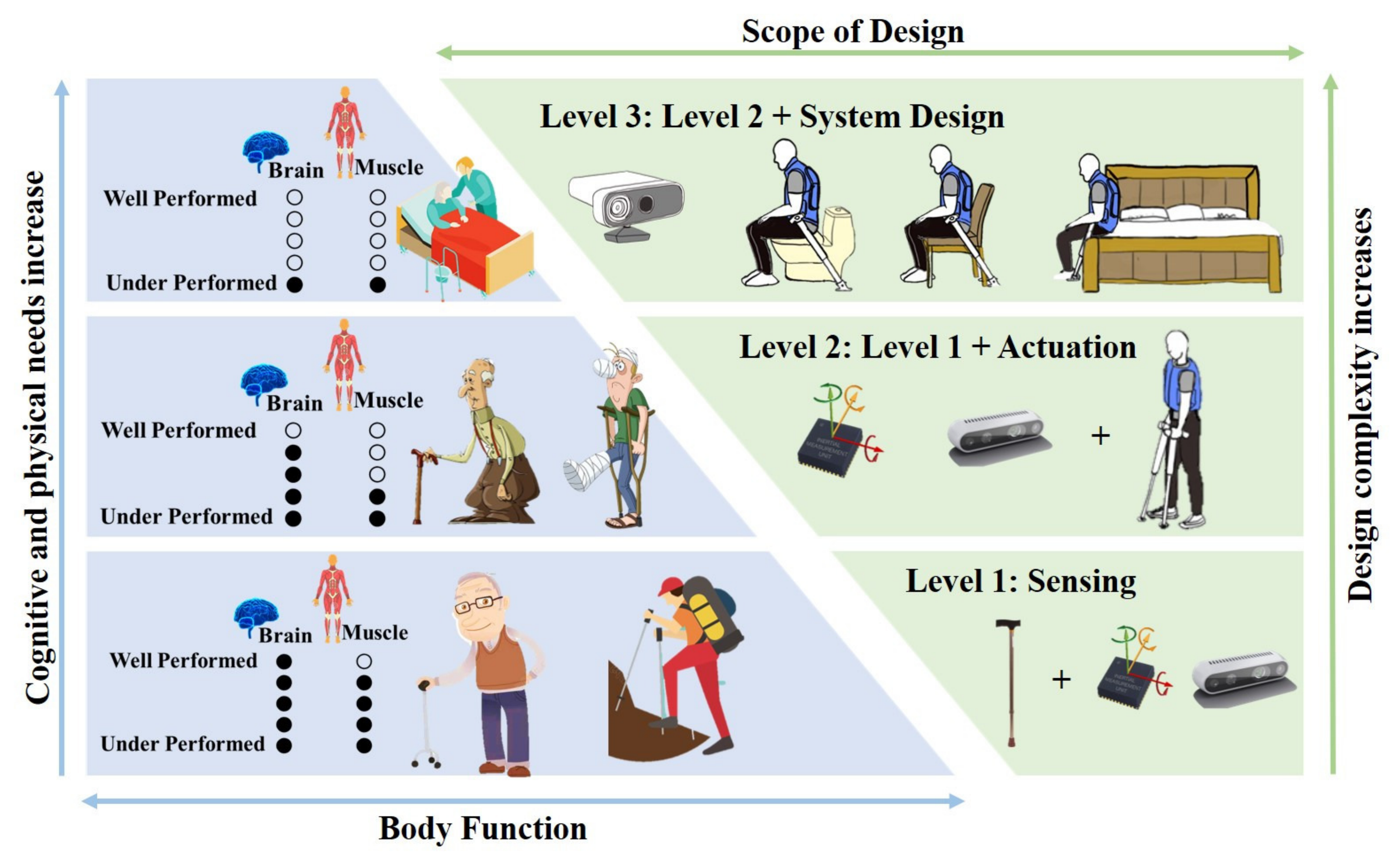}}
        \caption{Three-level design guidelines based on cognitive and physical conditions and corresponding design considerations. }
        \label{fig:ThreeLevelGuidlines}
    \end{figure}

    A critical question of the SuperLimbs is the communication between the human and extra limbs. As shown in Fig. \ref{fig:ThreeLevelGuidlines}, different brain-muscle conditions can be categorized into three classes. The corresponding demands for design are shown on the right. Here, the proposed robotic cane system is used as an example. For those whose brain performs well while the muscle undergoes minor degeneration, for instance, the healthy elderly and the climbers, a regular cane with environmental perceiving sensors is sufficient. The SuperLimbs of this level augments human capabilities, \cite{Gonzalez2018Design} is an excellent representative. For those who undergo a significant decline in physical condition while mild in cognitive, for instance, the elderly suffer from motor ability lost and the disabled, additional actuation besides sensors is necessary. The SuperLimbs of this level can help the user to regain normal abilities. For those who suffer from severe brain and muscle disable, a more advanced system with the fusion of multi-modality sensors is significant. The SuperLimbs of this level compensates both cognitive and physical capacities. 
  
    Based on different requirements, different levels of brain and muscle function can be remixed in other ways. We focus on the elderly of the second-level while providing an integrated system of the highest level for the entire home assistance in our proposed system. 
    
\section{Conclusion}
\label{sec:Conclusion}
    The soft robot system, which focuses on bed, chair, and toilet as three main places at home, aims at assisting the elderly during the sit-to-stand process. Three components are included in the design, the 1-DOF pneumatically-driven robotic cane, the inflatable vest as the soft human-robot interface, and the non-intrusive depth camera as an ambient sensor. Once the human intention is detected, the system will work automatically, inflating the vest and transmitting force in the stand-up direction, which is under real-time control. The proof-of-concept prototype performed well in compensation of ground reaction force during both the hip-lift-off and knee-hip extension phases according to the experiment. It eliminated the overshoot and results in relatively smooth and flat ground reaction force on feet, which reduces fall risks and requires fewer efforts for the elderly. Finally, we summarize a three-level design guideline, providing a reference to SuperLimbs as the assistive robot design corresponding to the demanding community with different brain-muscle conditions.
   
    However, some issues need to be further addressed and worked on, like power source, which would be either in portable size or fixed on a specific location, friction with the ground. Therefore, the next steps for the soft robot system include calibration and modification on both model and image identification algorithm, implementation of closed-loop control on the pneumatic cylinder, testing with more human users, especially the elderly users, and improvements on design details.
    
\bibliographystyle{IEEEtran}
\bibliography{IEEEexample}

\end{document}